\documentclass{article}
\usepackage{spconf,amsmath,graphicx, multirow, hyperref}

\DeclareMathOperator*{\argmax}{argmax} 

\title{PCLD: Point Cloud Layerwise Diffusion for Adversarial Purification}
\name{Mert Gülşen$^{\star}$ \qquad Batuhan Cengiz$^{\star}$ \qquad Yusuf H. Şahin $^{\dagger}$\qquad Gozde Unal$^{\star}$}
\address{$^{\star}$ Istanbul Technical University, Department of AI and Data Engineering, Istanbul, Turkey\\
 $^{\dagger}$Istanbul Technical University, Department of Computer Engineering, Istanbul, Turkey}

\begin{document}
%
\maketitle
\begin{abstract}
Point clouds are extensively employed in a variety of real-world applications such as robotics, autonomous driving and augmented reality. Despite the recent success of point cloud neural networks, especially for safety-critical tasks, it is essential to also ensure the robustness of the model. A typical way to assess a model's robustness is through adversarial attacks, where test-time examples are generated based on gradients to deceive the model. While many different defense mechanisms are studied in 2D, studies on 3D point clouds have been relatively limited in the academic field. Inspired from PointDP, which denoises the network inputs by diffusion, we propose \textbf{P}oint \textbf{C}loud \textbf{L}ayerwise 
 \textbf{D}iffusion (PCLD), a layerwise diffusion based 3D point cloud defense strategy. Unlike PointDP, we propagated the diffusion denoising after each layer to incrementally enhance the results. We apply our defense method to different types of commonly used point cloud models and adversarial attacks to evaluate its robustness. Our experiments demonstrate that the proposed defense method achieved results that are comparable to or surpass those of existing methodologies, establishing robustness through a novel technique. Code is available at \href{https://github.com/batuceng/diffusion-layer-robustness-pc}{https://github.com/batuceng/diffusion-layer-robustness-pc}.
\end{abstract}

\begin{figure}
    \centering
    \includegraphics[width=0.9\linewidth]{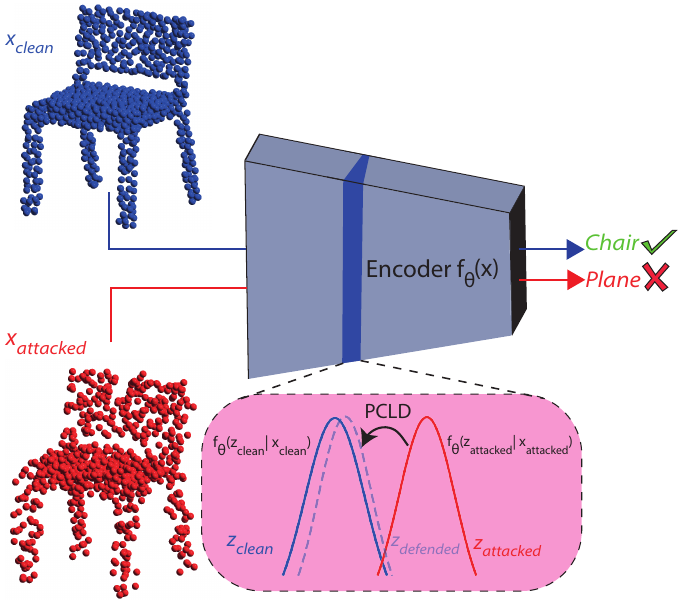}
    \caption{Overview of PCLD. In PCLD, the main focus is denoising the adversarial layer features back into the clean layer features with a diffusion-based purification.}
    \label{fig:1}
\end{figure}

\section{Introduction}
\label{sec:intro}

As with the 2D images, deep neural networks have markedly demonstrated their efficacy also for 3D point clouds by widely being used for many tasks like classification and segmentation \cite{Qi_2017_CVPR, klokov2017escape,zhou2023fat, li2023laptran}, registration \cite{deng2018ppf, mei2022partial} and scene reconstruction \cite{campagnolo2023fully, chen2019deep} in recent years. As a result of these methods and the significant impact of the usage of LiDAR technologies, point cloud deep neural networks has begun to be frequently used in safety-critical applications like autonomous driving \cite{chen2019pct, aygun20214d, cao2019adversarial} and industrial automation \cite{wang2023high}.

Despite the vulnerabilities of 2D deep learning methods being broadly investigated \cite{qiu2019review} and many defense mechanisms are suggested \cite{madry2017towards, nie2022diffusion}, the study of defenses for 3D point clouds is considerably less extensive. Similar to their 2D counterparts, adversarial attacks are a widespread application used to evaluate the robustness of 3D networks. These attacks, which can be in the form of black box \cite{wicker2019robustness} or white box attacks \cite{xiang2019generating, sun2021adversarially}, generate adversarial examples based on the network. Adversarial examples generally look similar to the human eye but are misclassified by the network. Instead of changing pixel values as in 2D, adversarial attacks in 3D focus on changing positions and counts of the points in a given sample.

Many defense mechanisms like DUP-Net \cite{zhou2019dup}, IF-Defense \cite{wu2020if} and PointDP aimed to purify the point cloud by cleaning the adversarial noise at the input level. On the other hand, the only solution to deal with the adversarial samples in feature domain is CCN \cite{li2022improving} which is a novel neural network having denoiser blocks after each convolutional layer. However, CCN cannot be extended for other classifier architectures as a defense method. 
Inspired by the aforementioned studies, we suggest \textbf{PCLD}, a diffusion-based layerwise purifier algorithm that can operate on high dimensional data without the hassle of retraining the neural network. We have demonstrated the overview of PCLD in Fig. \ref{fig:1} and explained it in detail later in Method section. Experiment results on the defenses of widely used point cloud classification networks like PointNet\cite{Qi_2017_CVPR}, DGCNN\cite{wang2019dynamic}, and PCT\cite{chen2019pct} showed that the proposed defense method achieved results that are comparable to or surpass those of existing methodologies.




\section{Related Work}
\textbf{Adversarial attacks.}
Point cloud based deep learning models have shown great success and continue to improve \cite{Qi_2017_CVPR,  wang2019dynamic, qi2017pointnet++, guo2021pct, ma2022rethinking, curvenet}. Despite their success, they have been discovered to be vulnerable against carefully crafted adversarial examples \cite{szegedy2013intriguing}. These adversarial data examples are indistinguishable from clean data examples to human perception, yet they are proven to completely degrade the accuracy of deep learning models. There exist numerous adversarial attack methods for creating adversarial examples. Goodfellow \textit{et al.} \cite{goodfellow2014explaining} proposed the Fast Gradient Sign Method (FGSM), which perturbs the input image in a single step toward the gradient that maximizes the loss, thereby creating an adversarial image. Kurakin \textit{et al.} \cite{kurakin2018adversarial} improved this idea by using multiple steps of FGSM to perform an adversarial attack. Madry \textit{et al.} \cite{madry2017towards} demonstrated Projected Gradient Descent (PGD) attack which limits the iterative gradient steps with a box constraint. Carlini \textit{et al.} \cite{carlini2017towards} proposed optimization-based C\&W attacks. 

The success of adversarial attacks on 2D models leads to studies adapting adversarial attacks to 3D deep learning models. Xiang \textit{et al.} \cite{xiang2019generating} were the first to demonstrate that 3D point cloud models are also vulnerable to adversarial attacks. They presented point perturbation and generation attacks that utilize the C\&W attack method to perturb points. Many of the following studies proposed the adaptation of different 2D adversarial attacks to 3D point cloud models: Liu \textit{et al.} \cite{liu2019extending} extended FGSM and Iterative FGSM (IFGSM) methods to 3D point cloud models. Zheng \textit{et al.} \cite{zheng2019pointcloud} used point cloud saliency maps for applying a point dropping attack. Sun \textit{et al.} \cite{sun2021adversarially} applied PGD attack to 3D point cloud models. Tsai \textit{et al.} \cite{tsai2020robust} proposed K-Nearest-Neighbor (kNN) attack, which adds a loss term to C\&W attack method using kNN distance for limiting distances between adjacent points. \\

\noindent \textbf{Adversarial Defenses.} Success of adversarial attacks on deep learning models raise the importance of developing defense methods due to many safety-critical applications \cite{Tu_2020_CVPR, cao2019adversarial}. Adversarial training \cite{goodfellow2014explaining, madry2017towards} is one of the possible countermeasures against adversarial attacks. This method uses both clean and adversarial examples to train a more robust model. Liu \textit{et al.} \cite{liu2019extending} extended adversarial training procedure to 3D point cloud models. Zhang \textit{et al.} \cite{zhang2022pointcutmix} proposed PointCutMix, which is a data augmentation method to improve robustness during training process. 

Beyond adversarial training and data augmentation, there exists adversarial purification \cite{shi2021online, nie2022diffusion} as another defense method, which aims to increase model robustness by readjusting adversarial inputs to align more closely with the true distribution, employing a range of transformation methods. \textit{Nie et al.} \cite{nie2022diffusion} proposed DiffPure, which uses Denoising Diffusion Probabilistic Models (DDPM) \cite{ho2020denoising} for adversarial purification as a preprocessing step on 2D inputs for classification. For 3D point clouds, \textit{Zhou et al.} \cite{zhou2019dup} proposed Denoising and UPsampler Network (DUP-Net) that uses statistical outlier removal (SOR) \cite{rusu2008towards} and a point upsampler network \cite{yu2018pu} for removing outliers and increasing surface smoothness on adversarial input point clouds. Furthermore, \textit{Wu et al.} \cite{wu2020if} proposed IF-Defense, which also uses SOR as first step and an implicit function network \cite{mescheder2019occupancy} for recovering surface of point clouds in the next steps to defend against 3D adversarial attacks. PointDP \cite{sun2023critical} extended 2D adversarial purification diffusion models to 3D, where a conditional 3D point cloud diffusion model is leveraged.

\section{Method}


In the preliminaries section, we will first define the basics of a classifier neural network with multiple layers and diffusion processes. After that, we will explain diffusion purification and our suggested method, \textbf{PCLD}, which suggests multi-layer diffusion purification for point cloud classification.

\subsection{Preliminaries}
\textbf{Classifiers.} For any input $x$ from class $y$, we can define a classifier $f_\theta(x)$ with $N$ layers as:

\vspace{-15pt}
\begin{gather} 
\label{eq:input} z^{(0)}=x \\ 
\label{eq:encoder} z^{(i+1)} = f^{(i)}_\theta(z^{(i)}), 0\leq i < N \\ 
\label{eq:cls_head} \hat{y} = f_\theta^{(N)}(z^{(N)})
\end{gather}

\vspace{-5pt}

We first define the layer-wise latent representation $z^{(i)}$ for values $i \in \{0,..,N\}$ where the initial $z^{(0)}$ is equal to the input, as shown in the Eq. \ref{eq:input}. In Eq. \ref{eq:encoder}, latent representation for the subsequent layer is calculated using the non-linear intermediate layer functions, $f_\theta^{(i)}$, which are parameterized over $\theta$. After that, the prediction $\hat{y}$ is calculated by the classification head $f^{(N)}$ in Eq. \ref{eq:cls_head}. We can define a loss function (e.g. cross-entropy loss) to calculate the error between true label $y$ and the prediction $\hat{y}$ as,

\begin{equation} \label{eq:crossentropy}
L(y, \hat{y}) = - \sum_{k} y_k \log(\hat{y}_k).
\end{equation}

\begin{figure*}
    \centering
    \includegraphics[width=\linewidth]{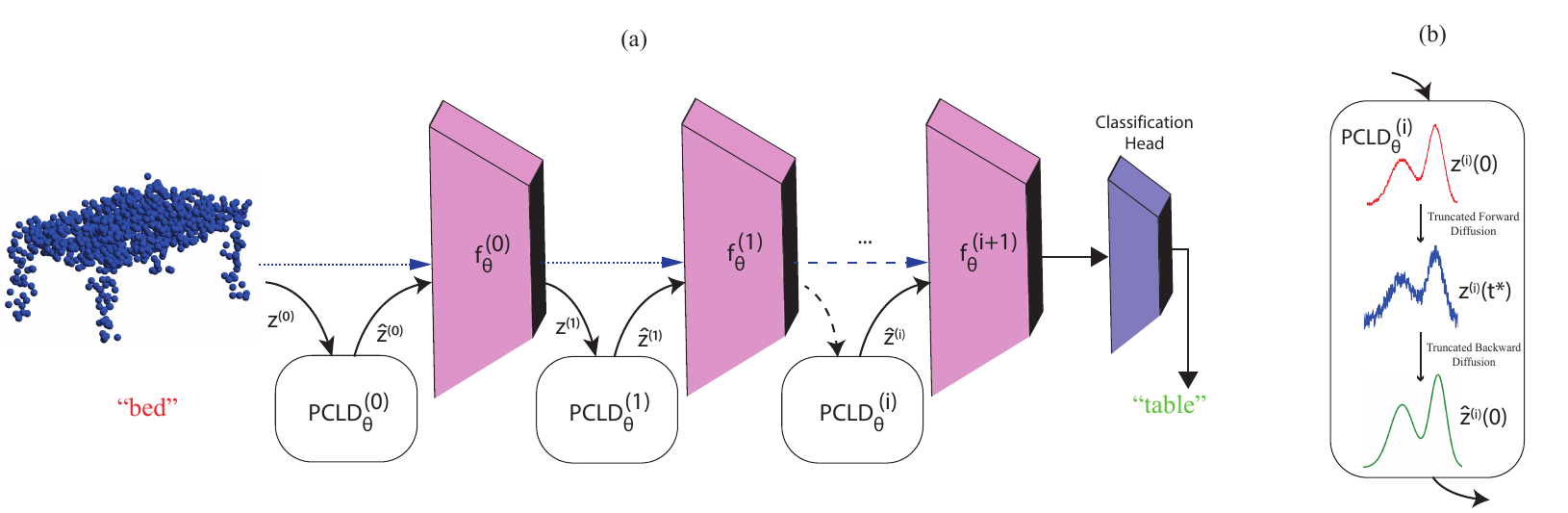}
    \caption{Overview of the \textbf{Point Cloud Layerwise Diffusion} method. The application of PCLD blocks in intermediate layers is illustrated in part (a), while the truncated diffusion process and details of the PCLD block are given in part (b).}
    \label{fig:method}
\end{figure*}

\noindent \textbf{Adversarial attacks.} An adversarial example is a sample that is within the $\epsilon$ neighbourhood $\mathcal{V}_\epsilon(x)$ of the input, which maximizes the loss function \cite{goodfellow2014explaining},

\begin{equation} \label{eq:adv_sample}
    x_{adv} = \argmax_{x \in \mathcal{V}_\epsilon(z^{(0)})} L(y, f_\theta(x)).
\end{equation}

The imperceptible small perturbation between $x_{adv}$ and $x$ in the input space, increases the output loss and causes incorrect predictions in the classifier network. 


\textbf{Assumption 4.1.} \textit{
An adversarial sample $x_{adv}$ that is within the $\epsilon$ distance to $x$ in the input layer $z^{(0)}$, gets farther from its unattacked location in every layer $z^{(i)}$ to maximize classification loss.
}\\

\noindent \textbf{Diffusion Probabilistic Models.} Consistent with the work by Luo and Hu \cite{Luo21DiffusionPC}, the diffusion process begins with a point cloud $x_0$ sampled from an unknown data distribution $q(x)$. The forward phase of the diffusion model gradually introduces Gaussian noise to this initial point cloud. This progression is mathematically represented as:

\begin{eqnarray}
q(x_{1:N} | x_0) := \prod_{n=1}^{N} q(x_n | x_{n-1})_{\sim \mathcal{N}(x_n; (1 - \beta_n)x_{n-1}, \beta_n I)}
\end{eqnarray}

\noindent where $x_n$ represents the $n^{th}$ step result generated within the process and $\beta_n$ is the noise scheduler which increases the noise variance progressively.

The reverse process aims to reconstruct the point cloud by progressively removing the added Gaussian noise using the equation,

\begin{eqnarray} \label{backdiff}
&& p_{\theta}(x_{0:N} | z_x) := \\ \nonumber
&& p(x_N) \prod_{n=1}^{N} p_{\theta}(x_{n-1} | x_n, z_x)_ {\sim \mathcal{N}(x_{n-1} | \mu_{\theta}(x_n, n, z_x), \beta_n I)} 
\end{eqnarray}

\noindent where $z_x$ is an encoded embedding for the diffusion guidance and $\mu_{\theta}$ is the mean value of the underlying distribution approximated by the diffusion network sampled as,  

\begin{gather}
\mu_{\theta}(x_n, n, z_x) = \frac{1}{\sqrt{1 - \beta_n}} \left( x_n - \beta_n \sqrt{1 - \alpha_n} \epsilon_{\theta}(x_n, n, z_x) \right), \nonumber \\
\alpha_n = \prod_{i=1}^{n} (1 - \beta_i) \label{eq:reparam_trick}
\end{gather}

\noindent following the original DDPM setup \cite{ho2020denoising} and the reparametrization trick. Consequently, considering both forward and backward processes, we end up with the following assumption.\\ 

\textbf{Assumption 4.2.} \textit{ \label{prob42}
A diffusion probabilistic model learns underlying data distrubition and it generates new samples by pushing noisy samples back into true distrubition in the reverse process with high precision.
}

\subsection{Point Cloud Layerwise Diffusion (PCLD)}

\textbf{Diffusion Purification.} As suggested in DiffPure for 2D \cite{nie2022diffusion} and PointDP for 3D \cite{pointdp23}, it is possible to purify adversarial noise from point clouds with forward and backward diffusion steps. Starting from $x_0 = x_{adv}$, the forward diffusion process for $t = (0,..,t^*)$ and $t^* \in (0, 1)$ can be computed as:


\begin{gather} \label{eq:fw}
x{(\frac{n}{N})} := x_n, \quad \beta{(\frac{n}{N})} := \beta_n, \quad \alpha{(\frac{n}{N})} := \alpha_n \nonumber \\
x(t^*) = \sqrt{\alpha(t^*)} x_{adv} + \sqrt{1 - \alpha(t^*)} \epsilon, \quad \epsilon \sim \mathcal{N}(0,I) 
\end{gather}

The purification result $\hat{x}(0)$ could be obtained using stochastic differential equation (SDE) solver $\textit{sdeint}(\cdot)$ \cite{nie2022diffusion, pointdp23} defined by the following equations:

\begin{gather}
\hat{x}(0) = \text{sdeint}(x(t^*), f_{\text{rev}}, g_{\text{rev}}, w, t^*, 0) \nonumber \\
f_{\text{rev}}(x, t, z_x) = -\frac{1}{2} \beta(t) [x + 2s_{\theta}(x, t, z_x)], \quad g_{\text{rev}}(t) = \sqrt{\beta(t)} \nonumber \\
s_{\theta}(x, t, z_x) = -\frac{1}{\sqrt{1 - \alpha(t)}} \epsilon_{\theta}(x(t), tN, z_x) \label{eq:bw}
\end{gather}

Investigating the difference between $x(0)$ and $\hat{x}(0)$, both PointDP and DiffPure pointed out that the adversarial noise in the given examples are eliminated during the diffusion process. 

\textbf{Assumption 4.3.} \textit{
Using a pretrained diffusion probabilistic model, adversarial noises can be purified with truncated number ($t^*<1$) of forward and backward steps. 
}\\

\textbf{PCLD: Point Cloud Layerwise Diffusion.} Inspired by PointDP, we expand the purification process to latent space where we replace the initial point cloud with the high dimensional layer data $x_0=z^{(i)}$. 

Based on our previous assumptions, we claim that, it is possible to train multiple layer-wise diffusion models to learn underlying distribution in every layer and hierarchically/recursively purify each layer to remove adversarial perturbations. We call our method \textbf{Point Cloud Layerwise Diffussion} (\textbf{PCLD}) and suggested the following training and inference methods.

In the training phase, we train a diffusion probabilistic model for the layer features $z^{(i)}$ of a pretrained classifier neural network $f_\theta(\cdot)$. We minimize the Fisher Divergence using the following loss:
\begin{equation}
\mathcal{L} = E_{z^{(i)}, t, \epsilon} [ \| \epsilon - \epsilon_{\theta} ( \sqrt{\alpha(t)} z^{(i)} + \sqrt{1 - \alpha(t)} \epsilon, t, e(z^{(i)}) ) \|^2 ]
\end{equation}

In inference however, we apply the truncated forward and backward steps following Eq. \ref{eq:fw_layer} \& \ref{eq:bw_layer}:

\begin{gather} 
z^{(i)}(t^*) = \sqrt{\alpha(t^*)} z^{(i)}_{adv} + \sqrt{1 - \alpha(t^*)} \epsilon, \quad \epsilon \sim \mathcal{N}(0,I) \label{eq:fw_layer} \\
\hat{z}^{(i)}(0) = \text{sdeint}(z^{(i)}(t^*), f_{\text{rev}}, g_{\text{rev}}, w, t^*, 0)  \label{eq:bw_layer}
\end{gather}

We have demonstrated the layerwise application processes of PCLD and the internal inference process in the Fig. \ref{fig:method} part (a) \& (b) respectively. Our suggested method can be applied to any trained classifier $f_\theta(\cdot)$ in a plug-and-play manner.

\section{Experiments}

\begin{table}[ht!]

\caption{Experiment results reported in terms of Accuracy (\%) percent. Each row correspond to a Model-attack pair while defense methods are given in the columns. The Clean attack correspondences to no attack being applied while defense is applied. Same applies for 'None' defend method. Best and second scores are shown by \textbf{bold} and \underline{underline} respectively.}
\centering
\label{results_table}
\scalebox{0.70}{
\begin{tabular}{cc||ccccccc|}
\cline{3-9}
\multicolumn{2}{c|}{}                                       & \multicolumn{7}{c|}{Defense}                                                                                                                                                                      \\ \hline
\multicolumn{1}{|c|}{\rotatebox[origin=c]{90}{ Model }}                       & \rotatebox[origin=c]{90}{Attacks} & \multicolumn{1}{c|}{\rotatebox[origin=c]{90}{None}}  & \multicolumn{1}{c|}{\rotatebox[origin=c]{90}{SRS}}   & \multicolumn{1}{c|}{\rotatebox[origin=c]{90}{SOR \cite{rusu2008towards}}}   & \multicolumn{1}{c|}{\rotatebox[origin=c]{90} {DUPNet \cite{zhou2019dup}}} & \multicolumn{1}{c|}{\rotatebox[origin=c]{90} { IF-Defense \cite{wu2020if} }} & \multicolumn{1}{c|}{\rotatebox[origin=c]{90} {PointDP \cite{sun2023critical}}} & \rotatebox[origin=c]{90} {PCLD \textbf{(ours)}} \\ \hline\hline

\multicolumn{1}{|c|}{\multirow{7}{*}{\rotatebox[origin=c]{90}{DGCNN \cite{wang2019dynamic}}}}      & Clean   & \multicolumn{1}{c|}{92.87} & \multicolumn{1}{c|}{84.81} & \multicolumn{1}{c|}{\textbf{91.65}} & \multicolumn{1}{c|}{67.18}  & \multicolumn{1}{c|}{86.06}      & \multicolumn{1}{c|}{ \begin{tabular}[c]{@{}l@{}}$88.19$\\ $\pm2.26$\end{tabular} }   & 

\begin{tabular}[c]{@{}l@{}}\underline{$89.10$}\\ \underline{$\pm1.66$}\end{tabular} 

\\ \cline{2-9} 
\multicolumn{1}{|c|}{}                            & Add     & \multicolumn{1}{c|}{63.29} & \multicolumn{1}{c|}{77.43} & \multicolumn{1}{c|}{77.39} & \multicolumn{1}{c|}{58.67}  & \multicolumn{1}{c|}{\underline{84.56}}      & \multicolumn{1}{c|}{83.14}   & \textbf{84.60}       \\ \cline{2-9} 
\multicolumn{1}{|c|}{}                            & CW      & \multicolumn{1}{c|}{0.28}  & \multicolumn{1}{c|}{61.83} & \multicolumn{1}{c|}{66.65} & \multicolumn{1}{c|}{36.75}  & \multicolumn{1}{c|}{82.86}      & \multicolumn{1}{c|}{\underline{85.78}}   & \textbf{86.47}       \\ \cline{2-9} 
\multicolumn{1}{|c|}{}                            & Drop    & \multicolumn{1}{c|}{72.20} & \multicolumn{1}{c|}{43.07} & \multicolumn{1}{c|}{\textbf{75.61}} & \multicolumn{1}{c|}{46.84}  & \multicolumn{1}{c|}{73.78}      & \multicolumn{1}{c|}{74.59}   & \underline{75.04}       \\ \cline{2-9} 
\multicolumn{1}{|c|}{}                            & KNN     & \multicolumn{1}{c|}{0.61}  & \multicolumn{1}{c|}{57.86} & \multicolumn{1}{c|}{61.18} & \multicolumn{1}{c|}{35.01}  & \multicolumn{1}{c|}{83.67}      & \multicolumn{1}{c|}{\underline{85.29}}   & \textbf{85.70}       \\ \cline{2-9} 
\multicolumn{1}{|c|}{}                            & PGD     & \multicolumn{1}{c|}{0.77}  & \multicolumn{1}{c|}{32.66} & \multicolumn{1}{c|}{20.62} & \multicolumn{1}{c|}{25.16}  & \multicolumn{1}{c|}{71.39}      & \multicolumn{1}{c|}{\underline{75.08}}   & \textbf{76.05}       \\ \cline{2-9} 
\multicolumn{1}{|c|}{}                            & PGDL2   & \multicolumn{1}{c|}{2.19}  & \multicolumn{1}{c|}{49.76} & \multicolumn{1}{c|}{46.03} & \multicolumn{1}{c|}{48.78}  & \multicolumn{1}{c|}{\textbf{81.93}}      & \multicolumn{1}{c|}{77.39}   & \underline{79.38}       \\ \hline\hline

\multicolumn{1}{|c|}{\multirow{7}{*}{\rotatebox[origin=c]{90}{PCT \cite{chen2019pct}}}}        & Clean   & \multicolumn{1}{c|}{92.95} & \multicolumn{1}{c|}{\underline{91.41}} & \multicolumn{1}{c|}{\textbf{92.42}} & \multicolumn{1}{c|}{85.53}  & \multicolumn{1}{c|}{89.47}      & \multicolumn{1}{c|}{\begin{tabular}[c]{@{}l@{}}$89.62$\\ $\pm3.34$\end{tabular} 
}   & 

\multicolumn{1}{c|}{\begin{tabular}[c]{@{}l@{}}$89.59$\\ $\pm1.15$\end{tabular} 
}
     \\ \cline{2-9} 
\multicolumn{1}{|c|}{}                            & Add     & \multicolumn{1}{c|}{63.09} & \multicolumn{1}{c|}{78.36} & \multicolumn{1}{c|}{80.15} & \multicolumn{1}{c|}{72.41}  & \multicolumn{1}{c|}{\textbf{86.14}}      & \multicolumn{1}{c|}{82.66}   & \underline{85.94}       \\ \cline{2-9} 
\multicolumn{1}{|c|}{}                            & CW      & \multicolumn{1}{c|}{0.00}  & \multicolumn{1}{c|}{80.79} & \multicolumn{1}{c|}{\textbf{89.71}} & \multicolumn{1}{c|}{77.35}  & \multicolumn{1}{c|}{\underline{88.86}}      & \multicolumn{1}{c|}{84.56}   & 87.52       \\ \cline{2-9} 
\multicolumn{1}{|c|}{}                            & Drop    & \multicolumn{1}{c|}{73.95} & \multicolumn{1}{c|}{73.91} & \multicolumn{1}{c|}{\underline{76.05}} & \multicolumn{1}{c|}{60.45}  & \multicolumn{1}{c|}{75.16}      & \multicolumn{1}{c|}{75.16}   & \textbf{77.88}       \\ \cline{2-9} 
\multicolumn{1}{|c|}{}                            & KNN     & \multicolumn{1}{c|}{0.65}  & \multicolumn{1}{c|}{76.42} & \multicolumn{1}{c|}{65.76} & \multicolumn{1}{c|}{65.80}  & \multicolumn{1}{c|}{\underline{86.02}}      & \multicolumn{1}{c|}{83.27}   & \textbf{86.10}       \\ \cline{2-9} 
\multicolumn{1}{|c|}{}                            & PGD     & \multicolumn{1}{c|}{3.16}  & \multicolumn{1}{c|}{45.42} & \multicolumn{1}{c|}{32.70} & \multicolumn{1}{c|}{36.51}  & \multicolumn{1}{c|}{\textbf{72.12}}      & \multicolumn{1}{c|}{55.55}   & \underline{69.33}       \\ \cline{2-9} 
\multicolumn{1}{|c|}{}                            & PGDL2   & \multicolumn{1}{c|}{9.81}  & \multicolumn{1}{c|}{44.85} & \multicolumn{1}{c|}{58.63} & \multicolumn{1}{c|}{64.67}  & \multicolumn{1}{c|}{\textbf{81.00}}      & \multicolumn{1}{c|}{62.03}   & \underline{73.70}       \\ \hline\hline

\multicolumn{1}{|c|}{\multirow{7}{*}{\rotatebox[origin=c]{90}{PointNet \cite{Qi_2017_CVPR}}}}   & Clean   & \multicolumn{1}{c|}{89.66}  & \multicolumn{1}{c|}{\underline{88.90}} & \multicolumn{1}{c|}{\textbf{89.34}} & \multicolumn{1}{c|}{88.37}  & \multicolumn{1}{c|}{86.35}      & \multicolumn{1}{c|}{

\begin{tabular}[c]{@{}l@{}}$88.16$\\ $\pm0.48$\end{tabular}
}   & 

\begin{tabular}[c]{@{}l@{}}$84.78$\\ $\pm3.11$\end{tabular}

\\ \cline{2-9} 
\multicolumn{1}{|c|}{}                            & Add     & \multicolumn{1}{c|}{51.34} & \multicolumn{1}{c|}{65.52} & \multicolumn{1}{c|}{77.59} & \multicolumn{1}{c|}{79.09}  & \multicolumn{1}{c|}{\textbf{84.68}}      & \multicolumn{1}{c|}{\underline{82.50}}   & 82.37       \\ \cline{2-9} 
\multicolumn{1}{|c|}{}                            & CW      & \multicolumn{1}{c|}{0.00}  & \multicolumn{1}{c|}{75.16} & \multicolumn{1}{c|}{86.35} & \multicolumn{1}{c|}{\textbf{86.51}}  & \multicolumn{1}{c|}{\underline{86.47}}      & \multicolumn{1}{c|}{85.25}   & 84.44       \\ \cline{2-9} 
\multicolumn{1}{|c|}{}                            & Drop    & \multicolumn{1}{c|}{45.30} & \multicolumn{1}{c|}{52.27} & \multicolumn{1}{c|}{51.82} & \multicolumn{1}{c|}{55.27}  & \multicolumn{1}{c|}{62.84}      & \multicolumn{1}{c|}{\textbf{71.03}}   & \underline{70.99}       \\ \cline{2-9} 
\multicolumn{1}{|c|}{}                            & KNN     & \multicolumn{1}{c|}{0.32}  & \multicolumn{1}{c|}{65.52} & \multicolumn{1}{c|}{70.95} & \multicolumn{1}{c|}{79.46}  & \multicolumn{1}{c|}{\underline{84.20}}      & \multicolumn{1}{c|}{\textbf{85.41}}   & 84.00       \\ \cline{2-9} 
\multicolumn{1}{|c|}{}                            & PGD     & \multicolumn{1}{c|}{4.09}  & \multicolumn{1}{c|}{24.07} & \multicolumn{1}{c|}{47.20} & \multicolumn{1}{c|}{60.29}  & \multicolumn{1}{c|}{\textbf{79.38}}      & \multicolumn{1}{c|}{70.42}   & \underline{72.08}       \\ \cline{2-9} 
\multicolumn{1}{|c|}{}                            & PGDL2   & \multicolumn{1}{c|}{0.00}  & \multicolumn{1}{c|}{6.20}  & \multicolumn{1}{c|}{48.30} & \multicolumn{1}{c|}{\underline{63.41}}  & \multicolumn{1}{c|}{\textbf{78.32}}      & \multicolumn{1}{c|}{33.67}   & 51.94       \\ \hline\hline

\multicolumn{1}{|c|}{\multirow{7}{*}{\rotatebox[origin=c]{90}{PointNet++ \cite{qi2017pointnet++}}}} & Clean   & \multicolumn{1}{c|}{91.05} & \multicolumn{1}{c|}{\underline{90.48}} & \multicolumn{1}{c|}{\textbf{90.96}} & \multicolumn{1}{c|}{87.40}  & \multicolumn{1}{c|}{87.76}      & \multicolumn{1}{c|}{

\begin{tabular}[c]{@{}l@{}}$86.92$\\ $\pm2.53$\end{tabular}

}   &

\begin{tabular}[c]{@{}l@{}}$86.99$\\ $\pm1.79$\end{tabular}

\\ \cline{2-9} 
\multicolumn{1}{|c|}{}                            & Add     & \multicolumn{1}{c|}{71.60} & \multicolumn{1}{c|}{78.69} & \multicolumn{1}{c|}{78.77} & \multicolumn{1}{c|}{77.03}  & \multicolumn{1}{c|}{\textbf{85.17}}      & \multicolumn{1}{c|}{82.25}   & \underline{82.98}       \\ \cline{2-9} 
\multicolumn{1}{|c|}{}                            & CW      & \multicolumn{1}{c|}{0.00}  & \multicolumn{1}{c|}{76.50} & \multicolumn{1}{c|}{\underline{85.01}} & \multicolumn{1}{c|}{82.50}  & \multicolumn{1}{c|}{\textbf{87.93}}      & \multicolumn{1}{c|}{80.83}   & 81.48       \\ \cline{2-9} 
\multicolumn{1}{|c|}{}                            & Drop    & \multicolumn{1}{c|}{\underline{81.60}} & \multicolumn{1}{c|}{80.31} & \multicolumn{1}{c|}{80.63} & \multicolumn{1}{c|}{75.00}  & \multicolumn{1}{c|}{79.25}      & \multicolumn{1}{c|}{81.60}   & \textbf{83.14}       \\ \cline{2-9} 
\multicolumn{1}{|c|}{}                            & KNN     & \multicolumn{1}{c|}{0.61}  & \multicolumn{1}{c|}{74.23} & \multicolumn{1}{c|}{62.40} & \multicolumn{1}{c|}{73.18}  & \multicolumn{1}{c|}{\textbf{86.35}}      & \multicolumn{1}{c|}{83.51}   & \underline{83.83}       \\ \cline{2-9} 
\multicolumn{1}{|c|}{}                            & PGD     & \multicolumn{1}{c|}{0.08}  & \multicolumn{1}{c|}{15.03} & \multicolumn{1}{c|}{7.21}  & \multicolumn{1}{c|}{28.89}  & \multicolumn{1}{c|}{\textbf{72.73}}      & \multicolumn{1}{c|}{67.83}   & \underline{67.99}       \\ \cline{2-9} 
\multicolumn{1}{|c|}{}                            & PGDL2   & \multicolumn{1}{c|}{1.22}  & \multicolumn{1}{c|}{34.81} & \multicolumn{1}{c|}{41.37} & \multicolumn{1}{c|}{64.83}  & \multicolumn{1}{c|}{\textbf{83.71}}      & \multicolumn{1}{c|}{72.61}   & \underline{73.14}       \\ \hline\hline

\multicolumn{1}{|c|}{\multirow{7}{*}{\rotatebox[origin=c]{90}{CurveNet\cite{curvenet}}}}   & Clean   & \multicolumn{1}{c|}{93.84} & \multicolumn{1}{c|}{88.53} & \multicolumn{1}{c|}{\textbf{91.13}}    & \multicolumn{1}{c|}{89.22}  & \multicolumn{1}{c|}{88.13}      & \multicolumn{1}{c|}{

\begin{tabular}[c]{@{}l@{}}$89.59$\\ $\pm2.42$\end{tabular}

}   & 

\begin{tabular}[c]{@{}l@{}}\underline{$90.93$}\\ \underline{$\pm2.10$}\end{tabular}

\\ \cline{2-9} 
\multicolumn{1}{|c|}{}                            & Add     & \multicolumn{1}{c|}{66.21} & \multicolumn{1}{c|}{74.84} & \multicolumn{1}{c|}{82.86}    & \multicolumn{1}{c|}{82.33}  & \multicolumn{1}{c|}{85.49}      & \multicolumn{1}{c|}{\underline{85.98}}   & \textbf{87.16}       \\ \cline{2-9} 
\multicolumn{1}{|c|}{}                            & CW      & \multicolumn{1}{c|}{0.00}  & \multicolumn{1}{c|}{72.33} & \multicolumn{1}{c|}{87.76}    & \multicolumn{1}{c|}{86.14}  & \multicolumn{1}{c|}{\underline{88.05}}      & \multicolumn{1}{c|}{85.45}   & \textbf{88.45}       \\ \cline{2-9} 
\multicolumn{1}{|c|}{}                            & Drop    & \multicolumn{1}{c|}{80.15} & \multicolumn{1}{c|}{55.19} & \multicolumn{1}{c|}{79.09}    & \multicolumn{1}{c|}{76.13}  & \multicolumn{1}{c|}{76.66}      & \multicolumn{1}{c|}{\underline{80.15}}   & \textbf{82.29}       \\ \cline{2-9} 
\multicolumn{1}{|c|}{}                            & KNN     & \multicolumn{1}{c|}{0.69}  & \multicolumn{1}{c|}{75.36} & \multicolumn{1}{c|}{71.68}    & \multicolumn{1}{c|}{82.50}  & \multicolumn{1}{c|}{\underline{86.47}}      & \multicolumn{1}{c|}{85.94}   & \textbf{88.57}       \\ \cline{2-9} 
\multicolumn{1}{|c|}{}                            & PGD     & \multicolumn{1}{c|}{4.09}  & \multicolumn{1}{c|}{30.83} & \multicolumn{1}{c|}{21.96}    & \multicolumn{1}{c|}{54.70}  & \multicolumn{1}{c|}{66.37}      & \multicolumn{1}{c|}{\underline{77.67}}   & \textbf{78.73}       \\ \cline{2-9} 
\multicolumn{1}{|c|}{}                            & PGDL2   & \multicolumn{1}{c|}{9.40}  & \multicolumn{1}{c|}{40.15} & \multicolumn{1}{c|}{53.00}    & \multicolumn{1}{c|}{71.76}  & \multicolumn{1}{c|}{\textbf{80.43}}      & \multicolumn{1}{c|}{74.88}   & \underline{76.86}       \\ \hline
\end{tabular}}
\end{table}

We have evaluated our proposed method with 5 different models and 6 different attacks on ModelNet40 \cite{modelnet40} dataset. Our setup consists of various pretrained classification neural networks: DGCNN \cite{wang2019dynamic}, PCT \cite{guo2021pct}, PointNet \cite{Qi_2017_CVPR}, PointNet++ \cite{qi2017pointnet++} and CurveNet \cite{curvenet}. PointNet and PointNet++ are the pioneering point cloud classification neural networks while DGCNN is a following work based on dynamic graphs. PCT and CurveNet on the other hand are more recent architectures with higher classification performance, based on attention and random walks.

For every classifier, we tested out the following white-box adversarial attacks: Add \cite{xiang2019generating}, C\&W \cite{carlini2017towards, xiang2019generating}, Drop \cite{zheng2019pointcloud}, kNN \cite{tsai2020robust}, PGD ($\|L_\infty\|$) and PGD-L2 ($\|L_2\|$) \cite{madry2017towards, sun2021adversarially} attacks. 

We compared our suggested method PCLD against multiple state-of-the-art adversarial purification methods. The compared methods are as follows: SRS \cite{zhou2019dup}, SOR \cite{rusu2008towards}, DUPNet \cite{zhou2019dup}, IF-Defense \cite{wu2020if} and PointDP \cite{pointdp23}. We selected the number of total steps in diffusion model as $N=200$. For finding the optimal truncated number of diffusion steps $t^*$, a grid search algorithm is used for both PointDP and our method PCLD.

\begin{table}[ht!]
\caption{List of truncated diffusion steps for PointDP and PCLD methods. Each row correspond to a Model-attack pair while corresponding layers are given in the columns. \\ }
\centering
\label{table:t_list}
\scalebox{0.8}{
\begin{tabular}{|cc||c||ccccc|}
\hline
\multicolumn{2}{|c|}{\multirow{2}{*}{}}              & PointDP     & \multicolumn{5}{c|}{PCLD}                                                                                                               \\ \cline{3-8} 
\multicolumn{2}{|c|}{}                               & \rotatebox[origin=c]{90}{ Input } & \multicolumn{1}{c|}{\rotatebox[origin=c]{90}{ Input }} & \multicolumn{1}{c|}{\rotatebox[origin=c]{90}{ Layer 1 }} & \multicolumn{1}{c|}{\rotatebox[origin=c]{90}{ Layer 2 }} & \multicolumn{1}{c|}{\rotatebox[origin=c]{90}{ Layer 3 }} & \rotatebox[origin=c]{90}{ Layer 4 } \\ \hline\hline
\multicolumn{1}{|c|}{\multirow{6}{*}{\rotatebox[origin=c]{90}{DGCNN}}} & Add   & 5           & \multicolumn{1}{c|}{5}           & \multicolumn{1}{c|}{15}      & \multicolumn{1}{c|}{20}      & \multicolumn{1}{c|}{5}       & 25      \\ \cline{2-8} 
\multicolumn{1}{|c|}{}                       & CW    & 10          & \multicolumn{1}{c|}{5}           & \multicolumn{1}{c|}{0}       & \multicolumn{1}{c|}{30}      & \multicolumn{1}{c|}{10}      & 40      \\ \cline{2-8} 
\multicolumn{1}{|c|}{}                       & Drop  & 5           & \multicolumn{1}{c|}{5}           & \multicolumn{1}{c|}{0}       & \multicolumn{1}{c|}{30}      & \multicolumn{1}{c|}{10}      & 40      \\ \cline{2-8} 
\multicolumn{1}{|c|}{}                       & kNN   & 10          & \multicolumn{1}{c|}{5}           & \multicolumn{1}{c|}{5}       & \multicolumn{1}{c|}{35}      & \multicolumn{1}{c|}{10}      & 50      \\ \cline{2-8} 
\multicolumn{1}{|c|}{}                       & PGD   & 15          & \multicolumn{1}{c|}{10}          & \multicolumn{1}{c|}{5}       & \multicolumn{1}{c|}{45}      & \multicolumn{1}{c|}{0}       & 70      \\ \cline{2-8} 
\multicolumn{1}{|c|}{}                       & PGDL2 & 25          & \multicolumn{1}{c|}{15}          & \multicolumn{1}{c|}{5}       & \multicolumn{1}{c|}{50}      & \multicolumn{1}{c|}{0}       & 25      \\ \hline\hline
\multicolumn{1}{|c|}{\multirow{6}{*}{\rotatebox[origin=c]{90}{PCT}}} & Add   & 5           & \multicolumn{1}{c|}{5}           & \multicolumn{1}{c|}{30}      & \multicolumn{1}{c|}{35}      & \multicolumn{1}{c|}{15}      & 85      \\ \cline{2-8} 
\multicolumn{1}{|c|}{}                       & CW    & 5           & \multicolumn{1}{c|}{5}           & \multicolumn{1}{c|}{10}      & \multicolumn{1}{c|}{60}      & \multicolumn{1}{c|}{35}      & 10      \\ \cline{2-8} 
\multicolumn{1}{|c|}{}                       & Drop  & 5           & \multicolumn{1}{c|}{5}           & \multicolumn{1}{c|}{15}      & \multicolumn{1}{c|}{90}      & \multicolumn{1}{c|}{15}      & 55      \\ \cline{2-8} 
\multicolumn{1}{|c|}{}                       & kNN   & 5           & \multicolumn{1}{c|}{5}           & \multicolumn{1}{c|}{25}      & \multicolumn{1}{c|}{50}      & \multicolumn{1}{c|}{0}       & 100     \\ \cline{2-8} 
\multicolumn{1}{|c|}{}                       & PGD   & 10          & \multicolumn{1}{c|}{5}           & \multicolumn{1}{c|}{35}      & \multicolumn{1}{c|}{85}      & \multicolumn{1}{c|}{70}      & 30      \\ \cline{2-8} 
\multicolumn{1}{|c|}{}                       & PGDL2 & 10          & \multicolumn{1}{c|}{5}           & \multicolumn{1}{c|}{35}      & \multicolumn{1}{c|}{100}     & \multicolumn{1}{c|}{60}      & 0       \\ \hline\hline
\multicolumn{1}{|c|}{\multirow{6}{*}{\rotatebox[origin=c]{90}{PointNet}}} & Add   & 10          & \multicolumn{1}{c|}{15}          & \multicolumn{1}{c|}{25}      & \multicolumn{1}{c|}{10}      & \multicolumn{1}{c|}{0}       & -       \\ \cline{2-8} 
\multicolumn{1}{|c|}{}                       & CW    & 30          & \multicolumn{1}{c|}{15}          & \multicolumn{1}{c|}{15}      & \multicolumn{1}{c|}{15}      & \multicolumn{1}{c|}{5}       & -       \\ \cline{2-8} 
\multicolumn{1}{|c|}{}                       & Drop  & 80          & \multicolumn{1}{c|}{70}          & \multicolumn{1}{c|}{30}      & \multicolumn{1}{c|}{15}      & \multicolumn{1}{c|}{0}       & -       \\ \cline{2-8} 
\multicolumn{1}{|c|}{}                       & kNN   & 30          & \multicolumn{1}{c|}{15}          & \multicolumn{1}{c|}{15}      & \multicolumn{1}{c|}{15}      & \multicolumn{1}{c|}{5}       & -       \\ \cline{2-8} 
\multicolumn{1}{|c|}{}                       & PGD   & 30          & \multicolumn{1}{c|}{25}          & \multicolumn{1}{c|}{25}      & \multicolumn{1}{c|}{0}       & \multicolumn{1}{c|}{0}       & -       \\ \cline{2-8} 
\multicolumn{1}{|c|}{}                       & PGDL2 & 30          & \multicolumn{1}{c|}{15}          & \multicolumn{1}{c|}{70}      & \multicolumn{1}{c|}{35}      & \multicolumn{1}{c|}{0}       & -       \\ \hline\hline
\multicolumn{1}{|c|}{\multirow{6}{*}{\rotatebox[origin=c]{90}{PointNet++}}} & Add   & 5           & \multicolumn{1}{c|}{5}           & \multicolumn{1}{c|}{15}      & \multicolumn{1}{c|}{20}      & \multicolumn{1}{c|}{0}       & -       \\ \cline{2-8} 
\multicolumn{1}{|c|}{}                       & CW    & 15          & \multicolumn{1}{c|}{10}          & \multicolumn{1}{c|}{35}      & \multicolumn{1}{c|}{30}      & \multicolumn{1}{c|}{0}       & -       \\ \cline{2-8} 
\multicolumn{1}{|c|}{}                       & Drop  & 0           & \multicolumn{1}{c|}{0}           & \multicolumn{1}{c|}{45}      & \multicolumn{1}{c|}{80}      & \multicolumn{1}{c|}{0}       & -       \\ \cline{2-8} 
\multicolumn{1}{|c|}{}                       & kNN   & 15          & \multicolumn{1}{c|}{10}          & \multicolumn{1}{c|}{15}      & \multicolumn{1}{c|}{10}      & \multicolumn{1}{c|}{0}       & -       \\ \cline{2-8} 
\multicolumn{1}{|c|}{}                       & PGD   & 30          & \multicolumn{1}{c|}{20}          & \multicolumn{1}{c|}{30}      & \multicolumn{1}{c|}{25}      & \multicolumn{1}{c|}{0}       & -       \\ \cline{2-8} 
\multicolumn{1}{|c|}{}                       & PGDL2 & 45          & \multicolumn{1}{c|}{30}          & \multicolumn{1}{c|}{5}       & \multicolumn{1}{c|}{0}       & \multicolumn{1}{c|}{0}       & -       \\ \hline\hline
\multicolumn{1}{|c|}{\multirow{6}{*}{\rotatebox[origin=c]{90}{CurveNet}}} & Add   & 10          & \multicolumn{1}{c|}{5}           & \multicolumn{1}{c|}{90}      & \multicolumn{1}{c|}{20}      & \multicolumn{1}{c|}{50}      & 20      \\ \cline{2-8} 
\multicolumn{1}{|c|}{}                       & CW    & 20          & \multicolumn{1}{c|}{5}           & \multicolumn{1}{c|}{100}     & \multicolumn{1}{c|}{55}      & \multicolumn{1}{c|}{10}      & 0       \\ \cline{2-8} 
\multicolumn{1}{|c|}{}                       & Drop  & 0           & \multicolumn{1}{c|}{0}           & \multicolumn{1}{c|}{60}      & \multicolumn{1}{c|}{30}      & \multicolumn{1}{c|}{25}      & 100     \\ \cline{2-8} 
\multicolumn{1}{|c|}{}                       & kNN   & 20          & \multicolumn{1}{c|}{5}           & \multicolumn{1}{c|}{95}      & \multicolumn{1}{c|}{55}      & \multicolumn{1}{c|}{40}      & 45      \\ \cline{2-8} 
\multicolumn{1}{|c|}{}                       & PGD   & 95          & \multicolumn{1}{c|}{100}         & \multicolumn{1}{c|}{85}      & \multicolumn{1}{c|}{25}      & \multicolumn{1}{c|}{15}      & 0       \\ \cline{2-8} 
\multicolumn{1}{|c|}{}                       & PGDL2 & 30          & \multicolumn{1}{c|}{20}          & \multicolumn{1}{c|}{75}      & \multicolumn{1}{c|}{40}      & \multicolumn{1}{c|}{0}       & 55      \\ \hline
\end{tabular}}
\end{table}

We have showed our test outcomes in Table \ref{results_table}. The experimental findings indicate that our approach excels in 5 out of 6 attacks in the CurveNet Model and 4 out of 6 attacks in DGCNN, ranking second in the remaining cases. Regarding PCT and PointNet++, our model ranks within the top 2 defense methods in 5 out of 6 attacks. In the case of PointNet, our results are comparable to other defense methods. We propose that the reduced performance observed in PointNet and PointNet++ could be related to the translation network in the initial layer. Additionally, for the other three models, we rank second best on clean point clouds following the SOR. This is naturally expected, as the SOR method is specifically designed to eliminate outliers, which have minimal impact on clean samples.

In Table \ref{table:t_list}, we present the selected truncated number of diffusion steps $t^*$ for both PointDP and our method PCLD. Among the models, DGCNN, PCT, and CurveNet feature four purified layers, while PointNet and PointNet++ have three layers based on their architectures. Notably, for the PCT model, the number of diffusion steps taken at Input-Layer 1 is relatively low compared to Layers 2-4. When considering that our PCLD model consistently outperforms PointDP by at least 3\% accuracy in every attack scenario for PCT, the significance of purification in deeper layers becomes evident.

Overall, we surpass our pioneering diffusion based purification method PointDP and recieve comparable results against other defense methods.

\section{Discussion \& Conclusion}
In this paper, we introduced PCLD (Point Cloud Layerwise Diffusion), a novel defense strategy for enhancing the robustness of 3D point cloud classification models against adversarial attacks. Building upon the success of diffusion-based purification methods like PointDP, we extended the concept to operate on a layerwise basis within the neural network architecture. Our method involves training diffusion probabilistic models for each layer of a pretrained classifier, allowing us to hierarchically purify the features at multiple levels. Comparative analysis against existing defense mechanisms revealed that PCLD outperforms or achieves comparable results, underscoring its potential as a state-of-the-art defense strategy for 3D point cloud models. Particularly, we observed significant improvements in defenses against attacks on deeper layers, highlighting the importance of purification at multiple levels within the network architecture. 

Moreover, the adaptability of PCLD to different model architectures and attack scenarios suggests its potential applications across diverse domains, including safety-critical applications like autonomous driving and robotics. However, challenges such as dataset-specific effectiveness, computational overhead, and adaptation to advanced attacks remain, necessitating ongoing research for refinement and broader deployment of PCLD in real-world applications. Overall, our findings showcase the effectiveness of PCLD in enhancing the robustness of 3D point cloud classification models, paving the way for more resilient applications in safety-critical domains such as autonomous driving and robotics.

\bibliographystyle{IEEEbib}
\bibliography{refs}

\end{document}